\documentclass[runningheads]{llncs}
\usepackage{lipsum}  
\usepackage{appendix}
\usepackage{graphicx}
\usepackage{amsmath}
\usepackage{amssymb}

\usepackage{multirow} % for tables
\usepackage{array} % for tables
\usepackage[table,xcdraw,dvipsnames]{xcolor} % for tables and highlights
\usepackage{amssymb}% http://ctan.org/pkg/amssymb
\usepackage{pifont}% http://ctan.org/pkg/pifont
\usepackage{subcaption}% subfigure and subcaption

\definecolor{MyGray}{HTML}{EFEFEF}
\definecolor{MyOrange}{HTML}{FFCE93}

\newcommand{\cmark}{\ding{51}}%
\newcommand{\xmark}{\ding{55}}%

\usepackage{orcidlink} % for orcid

\begin{document}
\newcommand{\resnet}{ResNet-50}
\newcommand{\blip}{BLIP2-Qformer}
\newcommand{\vltfive}{VL-T5}
\newcommand{\textcloze}{\textit{text-cloze}}

\title{Multimodal Transformer for Comics Text-Cloze}
% Second Author\inst{2,3}\orcidID{1111-2222-3333-4444}
\author{Emanuele Vivoli \thanks{corresponding authors} \inst{1,2} \orcidlink{0000-0002-9971-8738} \and
Joan Lafuente Baeza $^{\tiny\star}$ \inst{1} \orcidlink{0009-0007-9736-9048} \and
Ernest Valveny Llobet \inst{1} \orcidlink{0000-0002-0368-9697} \and
Dimosthenis Karatzas \inst{1} \orcidlink{0000-0001-8762-4454}}
\authorrunning{E. Vivoli et al.}
\institute{CVC, Universitat Autonóma de Barcelona, Bellaterra 08193, Spain \and
MICC, University of Florence, Florence 50134, Italy \\
\email{evivoli@cvc.uab.cat} \\
\email{joan.lafuente@autonoma.cat}}
\maketitle              % typeset the header of the contribution
\begin{abstract}
This work explores a closure task in comics, a medium where visual and textual elements are intricately intertwined. Specifically, \textcloze\;refers to the task of selecting the correct text to use in a comic panel, given its neighbouring panels. Traditional methods based on recurrent neural networks, have struggled with this task due to limited OCR accuracy and inherent model limitations. We introduce a novel Multimodal Large Language Model (Multimodal-LLM) architecture, specifically designed for \textcloze, achieving a 10\% improvement over existing state-of-the-art models in both its easy and hard variants. Central to our approach is a Domain-Adapted ResNet-50 based visual encoder, fine-tuned to the comics domain in a self-supervised manner using SimCLR. This encoder delivers comparable results to more complex models with just one-fifth of the parameters. Additionally, we release new OCR annotations for this dataset, enhancing model input quality and resulting in another 1\% improvement. Finally, we extend the task to a generative format, establishing new baselines and expanding the research possibilities in the field of comics analysis.

\end{abstract}

\section{Introduction}
\iffalse
\begin{itemize}
    \item text-close task, interesting as visual and text both contribute (paper shows "On text cloze, accuracy increases when models are given images (in the form of pretrained VGG-16 features) in addition to text; on the other tasks, incorporating both modalities is less important.")
    \item hierarchical lstm
    \item flaw in sota:
    \begin{itemize}
        \item rnn is the only architecture used so far, we investigated whether we can improve:
        \item text quality (old OCR)
        \item image representation quality (VGG-16)
        \item network architecture
    \end{itemize}
    \item address the text-close task with transformer t5
    \item using t5 and improving both the image and text quality we surpassed sota results
    \item we performed exhaustive ablation studies to choose image and text features extractors
    \item CONTRIBUTIONS:
     \begin{itemize}
        \item We propose a new Multimodal-LLM based architecture for the comics \textcloze\;task
        \item We explore different image representations, and demonstrate that adapting a ResNet architecture in a self-supervised manner to the comics domain yields comparative results to newer Multimodal LLM image encoders
        \item We release new OCR data for the original dataset
        \item We define a new version of the task addressing generation
    \end{itemize}
\end{itemize}
\fi 

\begin{figure}[!h]
    \centering
    \includegraphics[width=0.95\textwidth]{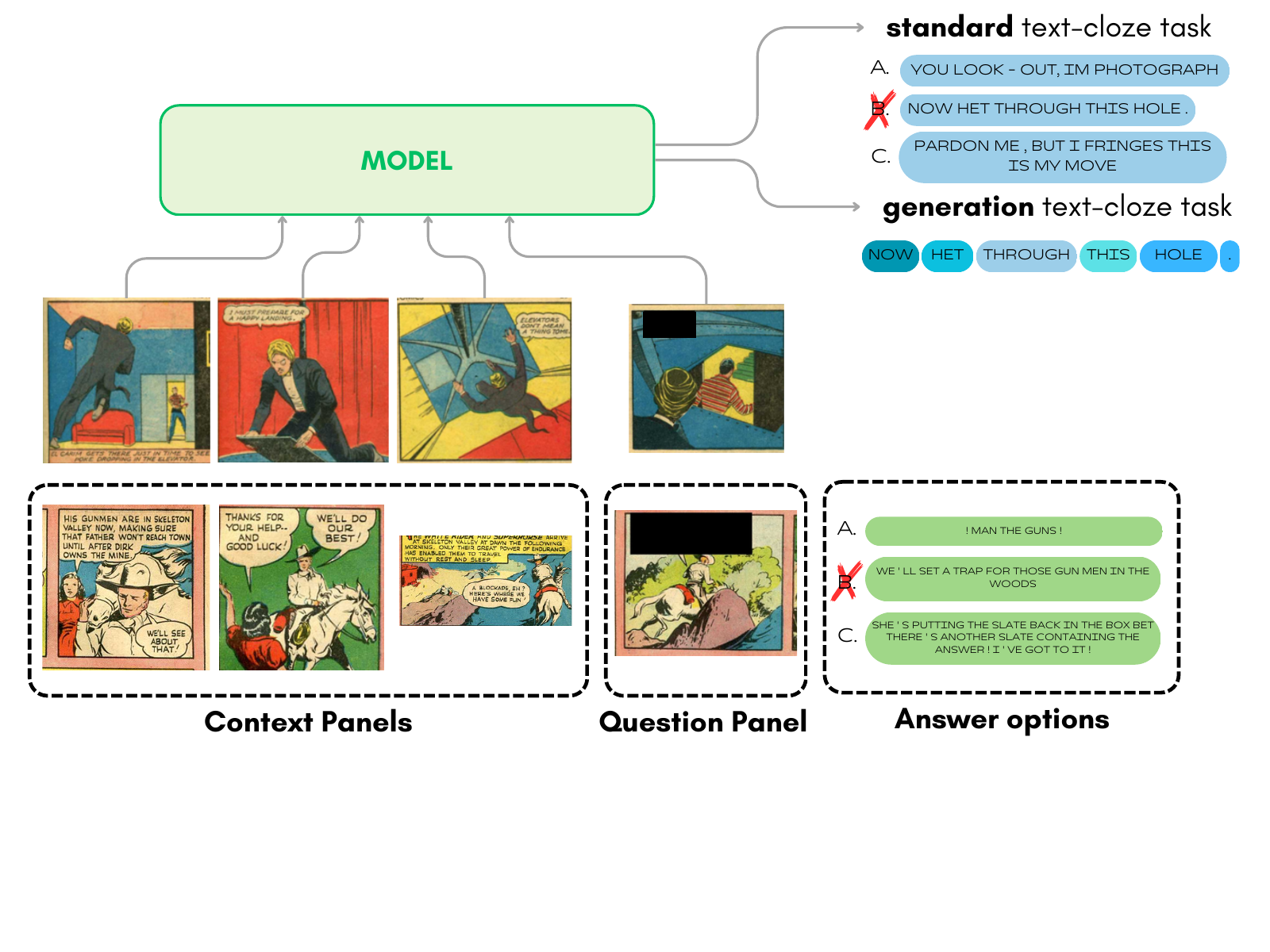}
    \vspace{-2cm}
    \caption{Two instances of the \textcloze\;task: a question panel (with masked balloons) is provided together with three candidate answers. Three context panels are provided to help choose the correct answer.}
    \label{fig:example_text_cloze}
    \vspace{-5mm}
\end{figure}

In the evolving landscape of multimodal data analysis, the study of comics presents a unique challenge, intertwining visual and textual elements. This fusion forms a distinct narrative structure, where the synergy between images and text is complex and integral. This is particularly evident in the \textcloze\;task, where models must predict missing text in a comic panel, given the panel with obscured text (question panel) and the previous three panels (context panels). The task is framed as a multiple-choice question, offering a number of distractor texts and the correct text originally on the panel (Figure \ref{fig:example_text_cloze}). 

Prior work in this area has mainly utilized recurrent neural networks \cite{iyyer_amazing_2017}. Soykan et al. \cite{soykan_comprehensive_2022} underscored the impact of OCR accuracy on model effectiveness, pointing out that subpar OCR transcriptions are a limiting factor. However, as we show in this work, advancements in OCR technology, while important, are insufficient by themselves to advance this task significantly. Considering this, progress also necessitates enhancements in the underlying model architecture. 

In this work, addressing the \textcloze\;task in comics, we make the following contributions:
\begin{enumerate}
    \item We introduce a novel Multimodal-LLM based architecture specifically designed for the comics \textcloze\;task, outperforming existing models by 10\% in both easy and hard variants of the task.
    \item Compare various image representations and demonstrate that finetuning \resnet\;to the domain of the comic in a self-supervised manner (SimCLR) yields comparable results to advanced Multimodal LLM image encoders, whilst having one-fifth of the parameters.
    \item We release new OCR data (Textract) for the original dataset.    
    \item We formulate a new version of the task addressing generation, marking a significant step forward in this area of research.
\end{enumerate}

\section{Related works}

\subsection{Comics Tasks in Vision and Language}
% second draft
Comics, as a medium, is thought to be multimodal from its inception. It consists of multiple panels on a page, each featuring text in balloons and recurrent characters with varying appearances. This complexity is amplified by animated objects and onomatopoeias, adding depth to the narrative. Among all datasets available \cite{guerin_ebdtheque_2013,fujimoto_manga109_2016,dunst_graphic_2017,soykan_comprehensive_2022,dutta_bcbid_2022}, the most common task is the detection of the various elements just described. The differences in style from one artist to another make difficult also the detection of text, a task that is largely explored in Computer Vision \cite{ye_text_2015,zhu_scene_2016,raisi_text_2020}. As so, many works in the field of Japanese comics (manga) propose tasks of onomatopoeia detection and their linking when an onomatopoeia is split into multiple pieces across the pages \cite{louis_detection_2023,louis_can_2023,baek_coo_2022}.

However, one important feature of comics is understanding the storyline, which cannot emerge from training detectors. The understanding has been designed as a collection of ``closure tasks'' in comics \cite{mccloud_understanding_1998}. These tasks are designed to probe the intricate relationship between text and images, given a context of 3 panels and the text in those. The key tasks include \textcloze, which focuses on identifying the appropriate text for a given image (Figure \ref{fig:example_text_cloze}); \textit{visual-cloze}, which involves selecting the matching image for a given text; and \textit{character-coherence}, which assesses the continuity of character actions across panels.

Our research focuses on the \textcloze\;task, characterized by its unique challenges. This task is observed to be demanding, as reflected in its performance scores. In terms of computational approaches, earlier LSTM-based methods \cite{iyyer_amazing_2017,baek_coo_2022} have shown limited success in this area. This underscores the task's complexity, particularly in integrating visual and textual elements in comics effectively.

\subsection{Multimodal Large Language Models}

Recent years have witnessed a surge in the prominence of Language Models, marked by significant advancements \cite{radford_improving_2018,radford_language_2019,raffel_exploring_2020,touvron_llama_2023,chiang_vicuna_2023,touvron_llama_2023a}. A pivotal development in this domain has been the T5 model \cite{raffel_exploring_2020}, which introduced a unified framework for diverse NLP tasks, employing a text-to-text format.

Transformers have introduced substantial progress in text processing, primarily by overcoming the limitations associated with long-term memory states in previous models \cite{vaswaniashish_attention_2017}. While speed enhancement is a notable benefit, the key contribution lies in their ability to handle long-range dependencies more effectively. In image processing, transformers have also shown remarkable capabilities, as demonstrated by the Vision Transformer (ViT) which surpasses conventional convolutional architectures in certain aspects of image analysis \cite{dosovitskiy_vit_2021}. The effectiveness of transformers in processing both textual and visual data has naturally extended their application to multimodal contexts. This progression has led to the development of various multimodal vision and language Large Language Models (LLMs) \cite{wang_git_2022,cho_unifying_2021,zhu_minigpt4_2023,chen_minigptv2_2023}. 
% This led to the birth of multimodal vision and language LLMs \cite{wang_git_2022,cho_unifying_2021,zhu_minigpt4_2023,chen_minigptv2_2023}

Building on these developments, our research applies the \vltfive\;model \cite{cho_unifying_2021} to the \textcloze\;task in comics, selected for its balanced encoder-decoder architecture. This choice addresses the limitations of encoder-only architectures, as explored in our ablation studies. While models like MiniGPT-4 \cite{zhu_minigpt4_2023}, combining a Vision Transformer (ViT) with a linear layer and a decoder-only LLM, were considered, their larger parameter scale prompted us to defer their use to future work. The \vltfive\;model, consisting of approximately 224M parameters (112M each in encoder and decoder), presents a more manageable scale compared to the combined parameter sizes of larger models such as the 1.2B-parameter ViT from BLIP2 \cite{li_blip2_2023} and the smallest 7B-parameter Vicuna \cite{chiang_vicuna_2023}. This approach allows us to maintain a focus on efficiency while tackling the unique challenges of the \textcloze\;task in a multimodal context.

\section{Method}
\begin{figure}
    \centering
    \includegraphics[width=0.9\textwidth]{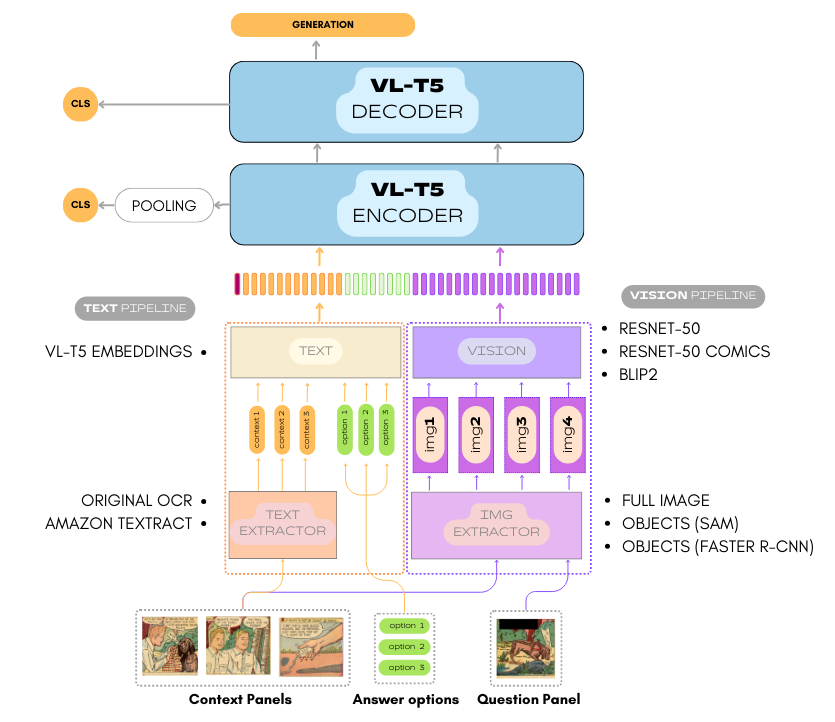}
    \caption{Architecture of \vltfive\;with text pipeline (OCR system extractor and \vltfive\;embeddings), and image pipeline (custom extractors and image encoders).}
    \label{fig:architecture}
    \vspace{-5mm}
\end{figure}

This section provides an overview of the fundamental elements pertaining to the architecture depicted in Figure \ref{fig:architecture}. It begins with a formal definition of the \textcloze\;task, followed by an explanation of the text and vision pipelines. Each of these pipelines is composed of two key modules: the extractor and the encoder.

\subsection{Task Definition}
Let $P = \{p_1, p_2, \cdots, p_n\}$ represent a sequence of comic panels, where each panel $p_i$ is composed of two modalities: the visual component $I_i$ referring to the image representation of a panel,
%, with any text content in it previously masked (We use the ground truth of the balloons to set the pixels to back, example in figure \ref{fig:example_mask})
and the text component $T_i$, which is the text that appears in the different balloons of the panel. $T_i$ comprehends the text up to three balloons, linearised using the order in which it is extracted by the OCR. The task is to predict the occluded text $T_{\text{target}}$ for a given panel $p_{\text{target}}$, utilizing the visual context $I_{target}^{\star} = M(I_{target})$ where $M$ is the masking process and $I_{target}^{\star}$ is the result of masking the text inside the panel image. For the task, three preceding panels are provided as context.
Finally, as for the original definition of the task \cite{iyyer_amazing_2017}, the probability distribution is also conditioned on the possible answer options $O = \{ O_0, O_1, \cdots, O_n\}$ which includes the target text $O_t$ and distractors text $O_j$ with $t,j \in [0, n]$, $j \neq t$, and $n$ is either $3$ or $10$, making it a multiple-choice question task. The final task consists of predicting the index $i$ of the correct answer and not the correct answer itself.
The final task is expressed by the following equation:
\begin{equation*}
    \hat{t} = \underset{i}{\mathrm{argmax}} \; P(T_i \mid O, I_{\text{target}}^{\star}, C)
\end{equation*}

If the task is tackled as a generation task, where tokens are generated one after the other, the previous formula can be expressed as:
\begin{equation*}
    T_{\text{predicted}} = \bigoplus_{k=1}^{K} \underset{T_k}{\mathrm{argmax}} \; P(T_k \mid T_{<k}, I_{\text{target}}^{\star}, C)
\end{equation*}
where $T_{\text{predicted}}$ is the generated text sequence, $T_k$ is the token generated at step $k$, $T_{<k}$ represents all previously generated tokens, and $K$ is the length of the generated sequence. The context $C$ includes the content from the three preceding panels. The operation $\bigoplus_{k=1}^{K}$ is used to denote the aggregation of individual tokens into a continuous sequence.

\subsection{Vision Pipeline}
As can be seen from Figure \ref{fig:architecture}, there are two blocks involved in creating the image representation (right pipeline, purple): the \textit{Image Extractor} and the \textit{Image Encoder}.

\subsubsection{Extractor.}
The Image Extractor module processes each panel image and employs one of two approaches: either retaining the original image in its entirety or applying region extraction to create a series of patches that depict the panel's contents. In the case of region extraction, these patches, representing various objects within the panel, collectively form a comprehensive representation of the panel. We use two different region extraction methods:

\paragraph{Faster R-CNN.} We utilize a convolutional-based object detector, Faster R-CNN \cite{ren_faster_2016}, for detecting and identifying objects in the panel images. Our implementation, adapted from the Detectron2 \cite{wu2019detectron2} repository, employs a ResNet-101 backbone trained on the Visual Genome dataset. This approach allows us to represent a panel with the most confidently detected objects. The rationale behind this object-focused representation stems from the intricate nature of comic scenes, where a mere image-based representation might fail to capture the full complexity of a comic panel.

\paragraph{Segment Anything Model (SAM).} As an alternative to the CNN-based method, we employ SAM, a transformer-based foundational model specifically designed for segmentation tasks \cite{kirillov_sam_2023}. SAM utilizes a ViT backbone along with a prompted masking light network for mask generation. Post mask generation, the segmented objects are extracted using their bounding boxes. The decision to use bounding boxes, as opposed to polygonal masks, is twofold: it maintains consistency and avoids potential alterations to object representations that could arise from cutting objects and filling backgrounds with solid colors, which might introduce unwanted noise. 

\paragraph{}
The details about Faster R-CNN and SAM hyperparameters are reported in Section \ref{sec:evaluation}.
The elements in Fig. \ref{fig:architecture}, denoted as [\textit{$\{img_1\}, \{img_2\}, \{img_3\}, \{img_4\}$}], are either the whole panel image or a collection of segmented elements from the panel with one of the approaches described before.

\subsubsection{Encoder.}
The Encoder works to extract features from images, which can be sourced from the entire panel or individual patches produced by the Extractor. In the case of panel-level representation, this process yields a single token that represents the entire image. Conversely, when representing the panel through objects, a token is produced for each object. Subsequently, these images are mapped into the \vltfive\;space. To facilitate this, a linear projection layer is employed, serving the essential purpose of aligning the output dimensions of the Encoder with the input dimensions required by the \vltfive\;model. We analyze several alternatives for the architecture of the encoder. 

% Following extraction, the Vision Encoder module processes each element using a neural network, transforming the image data -— whether it's a complete panel or an individual object -— into a feature representation that matches the dimensional input requirements of the \vltfive\; model. This step is essential as it transmutes visual information into a ``new word'', a token-like representation, that can be integrated within the \vltfive\;'s multimodal learning framework.

\paragraph{ResNet.} A \resnet\;backbone, pre-trained on ImageNet.

\paragraph{Comics ResNet.}
The \resnet\;model, initially trained on the ImageNet dataset, required adjustments to effectively process comic book imagery, due to the significant differences between these two visual domains. To bridge this gap, we adapted \resnet\;to better align with the unique features of comic images. This adaptation was essential, particularly in the absence of labeled comic datasets, leading us to adopt a self-supervised learning approach. Within this framework, we chose the SimCLR framework \cite{chen_simple_2020}, recognized for its efficacy in domain adaptation through contrastive learning. This approach was applied in two distinct ways: fine-tuning an existing ImageNet-trained \resnet\;model and training a new \resnet\;model from scratch.
The use of the SimCLR method is particularly apt given the large volume of varied, unlabeled comic data available. The primary goal is to enable the network to accurately recognize and understand the visual nuances of comic imagery. Although SimCLR can be applied to entire pages or panels, our focus was on the object level. This decision was made because defining similar objects is relatively easier, allowing the model to more effectively learn and represent objects with similar visual characteristics close to each other. Instead, the concept of what constitutes two similar pages or panels remains unclear when viewed at varying levels of granularity.
To this end, we have created a dataset of around 730k objects extracted from comic images using the Segment Anything Model (SAM), drawn from the first 250 books in our collection. This dataset has been instrumental in training the two versions of the \resnet\;model. This training approach enabled the models to learn a feature space meaningfully representing different comic elements, significantly enhancing their feature extraction capabilities. More details on our methodology and findings can be found in the supplementary materials.

\paragraph{\blip.} The Bidirectional Language-Image Pre-training for Vision Transformer, or BLIP2 \cite{li_blip2_2023}, is a model that has been pre-trained on a diverse set of images and associated textual descriptions. Its design is inherently multimodal, combining the strengths of Vision Transformers (ViT), and the new proposed Q-former, all with language supervision. This pretraining enables \blip\;to understand complex visual scenes and their narrative contexts, which is particularly advantageous for interpreting the rich visual language of comics.

\paragraph{}\noindent
In our study, we compare the above three encoders. \resnet, with less than a third of the parameters of \blip, serves as a reliable baseline. The fine-tuned \resnet\;aims to capture domain-specific nuances of comics, potentially matching the performance of more complex models. \blip, distinguished by its extensive multimodal pre-training, is included for its potential in interpreting complex visual narratives. This selection allows us to evaluate the impact of model size and domain-specific tuning on performance, as we will demonstrate in the experiments and results section.

\subsection{Text pipeline}
\subsubsection{Extractor and encoder.}
In our methodology, the text extraction process is fundamental, serving as the initial step for handling textual information. We utilize Optical Character Recognition (OCR) systems \cite{baek_character_2019,raisi_text_2020} to retrieve text from comic panels. These OCR systems are responsible for both detecting and recognizing text, and they additionally determine the sequence in which the text appears within each panel. OCR is instrumental in dealing with two types of text: contextual images (C) and multiple-choice answer options (O). For context images, the extracted text is used to enrich the text pipeline with explicit representation, complementing the visual data processed through the vision pipeline. 
We evaluated the efficacy of two OCR systems:

\paragraph{Original OCR}\cite{iyyer_amazing_2017}: For the initial task definition, text extraction was performed using an open-source OCR system. We have adopted these provided transcripts as a benchmark, enabling us to compare with the original methodologies and assess the impact of OCR quality by juxtaposing it with more advanced OCR technologies.

% \paragraph{ComicsText+ \cite{soykan_comprehensive_2022}:} An improved OCR system tailored to the distinctiveness of comic texts, which often includes stylized fonts and unconventional layouts. This system enhances the quality of the extracted text.

\paragraph{Amazon Textract} \protect\footnote{https://docs.aws.amazon.com/textract}: This is a high-precision, commercial OCR solution available through an API. Its adaptability for general purposes enables us to test its effectiveness in comic texts as well. As part of our contribution, we have generated new OCR transcriptions for every panel in our dataset, which we will make available for further research.

\begin{figure}
    \centering
    \includegraphics[width=\textwidth]{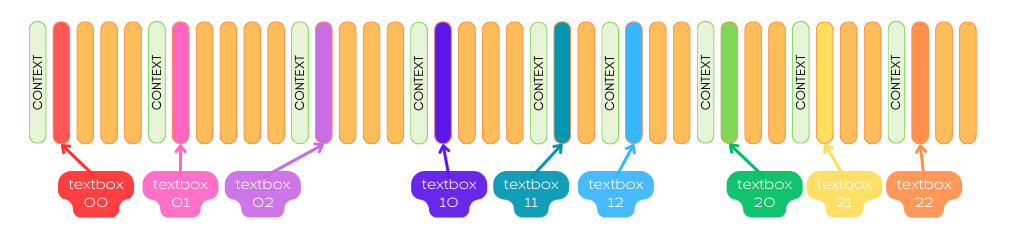}
    \caption{Representation of textboxes (balloon) separator tokens. Every group of tokens referred to textbox (balloon) has two tokens prefix that represents the context (light-green) and the textbox id (colored tokens). The orange tokens are the balloon text tokens.}
    \label{fig:text-separator-tokens}
    \vspace{-5mm}
\end{figure}

The process of extracting text involves identifying its position within each panel and grouping it with the corresponding speech balloon. After extracting text from the context panels, the \vltfive\;tokenizer breaks down the text into a series of individual tokens \( T = \{ t_1, t_2, \cdots, t_m\} \). These tokens are then assigned to specific embeddings via the embedding layer. Subsequently, tokens from context panels are combined and introduced with custom learnable tokens at the beginning. An example of these custom tokens is depicted in Figure \ref{fig:text-separator-tokens}. A similar approach is applied to the answer options.

\subsection{Proposed model}
We base our model on the \vltfive\;architecture \cite{cho_unifying_2021}, a multimodal transformer model that has been effectively employed in a variety of vision and language tasks \cite{ye_whether_2023,yang_crossing_2021}. \vltfive\;extends the T5 model to integrate visual data, making it suitable for multimodal tasks. In our adaptation of \vltfive\;for the comic \textcloze\;task, we maintain the core architecture but introduce task-specific modifications, and add the prefix-token \textcloze. Our implementation of \vltfive\;features two distinct configurations: a classification setup and a generation setup. In the classification setup, we either employ an encoder-only approach, where the encoder's final layer output is pooled into a single token for classification within a $[0, n]$ range (refer to ``pooling'' in Fig. \ref{fig:architecture}), or we use the full encoder-decoder model to generate the id of the selected option. The generation setup, on the other hand, prompts the model to produce the complete text of the option it deems correct.

\section{Experiments and Results}
Our model variants are built by combining different textual and visual extractors and encoders. A series of experiments are run to establish the best combination, as follows: %Our baselines are constructed as follows:
\begin{itemize}
    \item \textbf{Visual Extractors comparison.} Utilizing \vltfive\;with the original OCR system and various Image Extractors to establish which visual extractor works better to represent the comic panels.
    
    \item \textbf{Visual Encoders comparison.} After choosing the Image Extractor, benchmark the vision encoder components within the \vltfive\;framework, to explore the synergistic effects of multimodal inputs.
    
    \item \textbf{Textual comparison.} Employing \vltfive\;with different OCR systems using the same visual encoders, to gauge the impact of text quality on performance.
\end{itemize}

By employing these configurations, we aim to systematically dissect the contributions of each component within the \vltfive\;architecture. This methodology allows us to evaluate the choice of image extractor and encoder, the importance of high-quality OCR, and the overall effectiveness of the \vltfive\;model in integrating visual and textual information for the comic \textcloze\;task.

\subsection{Data}
Our experimental dataset, sourced from \cite{iyyer_amazing_2017}, encompasses the \textcloze\;task's training, validation, and test partitions across both easy and hard variants. It includes the OCR of the text within balloons, images of complete comic pages, and individual panels that have been automatically cropped. Furthermore, we have utilized a state-of-the-art OCR system to generate an updated OCR dataset, enhancing the text quality. Additionally, we have accurately mapped the panel coordinates corresponding to the cropped panels provided. A summarization of the dataset statistics used throughout our experiments is provided in Table \ref{tab:statistics_data}, offering an overview of the comprehensive nature of the data.

\begin{table}[h]
    \centering
    \vspace{-5mm}
    \caption{``Comics'' dataset (left) and \textcloze\;task (right) statistics.}
    \label{tab:statistics_data}
    \vspace{3mm}
    \begin{minipage}{.5\linewidth}
        \centering
        \begin{tabular}{lr}
            \multicolumn{2}{c}{Number of objects}    \\ \hline
            Books          & 3,948                   \\
            Pages          & 198,657                 \\
            Panels         & 1,229,664               \\
            Textboxes \;     & 2,498,657               \\ \hline
        \end{tabular}
    \end{minipage}%
    \begin{minipage}{.5\linewidth}
        \centering
        \begin{tabular}{lr}
            \multicolumn{2}{c}{\textcloze\;instances} \\ \hline
            Train          & 142,952                 \\
            Validation     & 22,490                  \\
            Test           & 11,909                  \\ \hline
        \end{tabular}
    \end{minipage}
    \vspace{-8mm}
\end{table}

\subsubsection{Easy vs. Hard variants.}
The \textcloze\;task presents two distinct levels of difficulty: easy and hard. The distinction between these levels lies in the source of the distractor options. For the easy variant, distractors are drawn from a broad pool of balloon texts across the dataset, whereas for the hard variant, distractors are chosen from balloons on pages close to the target panel. This implies that the distractors are more likely to be contextually similar to the correct answer, thus demanding a more nuanced understanding of the context for accurate prediction. In our experiments, each model is evaluated in both test sets, regardless of the training data. We define tests as Independent and Identically Distributed (iid) when the training and testing difficulties match (easy with easy, hard with hard). Conversely, tests are categorized as Out Of Distribution (ood) when there is a cross-evaluation (training on easy, testing on hard, and vice versa).

\subsection{Impact of Panel Representation}
\label{sec:evaluation}
Our investigation into visual information extraction compares the efficacy of global image representation (entire panel) versus cropped object representation. We utilized two distinct neural networks for object detection in this comparison: Faster R-CNN (from Detectron2) and SAM (Segment Anything Model).

For object extraction with Faster R-CNN, we selected up to 10 objects per panel based on confidence scores, with a minimum confidence threshold of 0.2. With SAM, we limited the number to 36 higher stability score objects per panel when more were available. We adjusted SAM's default parameters to optimize mask proposal and aggregation, setting \textit{points\_per\_side} to 20, \textit{pred\_iou\_thresh} to 0.85, \textit{stability\_score\_thresh} to 0.90, and \textit{crop\_n\_layers} to 0.

\begin{table}[h]
\vspace{-5mm}
\centering
\caption{Performance comparison of different Image Extractors using the \vltfive\;model, reported in Accuracy (\%). White rows indicate full image (panel-level) representation, while \colorbox{MyGray}{gray rows} denote object-level representation.}
\label{tab:easy-hard-image-extractor}
\vspace{3mm}
\begin{tabular}{clc|cc|cc}
model specifics $\downarrow$ & \multicolumn{2}{r|}{training data  $\rightarrow$} & \multicolumn{2}{c|}{} & \multicolumn{2}{c}{} \\
\multicolumn{2}{c|}{\textbf{Image}} & \textbf{Text} & \multicolumn{2}{c|}{\multirow{-2}{*}{\textbf{easy}}} & \multicolumn{2}{c}{\multirow{-2}{*}{\textbf{hard}}} \\ \hline
\textbf{Extractor} & \multicolumn{1}{c|}{\textbf{Encoder}} & \textbf{\begin{tabular}[c]{@{}c@{}}Extractor\\ (OCR)\end{tabular}} & \textbf{\begin{tabular}[c]{@{}c@{}}easy\\ (iid)\end{tabular}} & \textbf{\begin{tabular}[c]{@{}c@{}}hard\\ (ood)\end{tabular}} & \textbf{\begin{tabular}[c]{@{}c@{}}easy\\ (ood)\end{tabular}} & \textbf{\begin{tabular}[c]{@{}c@{}}hard\\ (iid)\end{tabular}} \\ \hline
\rowcolor[HTML]{EFEFEF} 
Faster R-CNN & \multicolumn{1}{l|}{\cellcolor[HTML]{EFEFEF}CRN scratch} & original & 77.28 & 62.52 & 75.42 & 68.13 \\
\rowcolor[HTML]{EFEFEF} 
SAM & \multicolumn{1}{l|}{\cellcolor[HTML]{EFEFEF}CRN scratch} & original & 75.69 & 61.39 & 75.80 & 67.32 \\ 
\xmark & \multicolumn{1}{l|}{CRN scratch} & original & 78.55 & \cellcolor[HTML]{FFCE93}\textbf{64.84} & 77.00 & \cellcolor[HTML]{FFCE93}\textbf{69.91} \\
\hline
% \rowcolor[HTML]{EFEFEF} 
% Faster R-CNN & \multicolumn{1}{l|}{\cellcolor[HTML]{EFEFEF}CRN finetuned} & original & - & - & - & - \\
\rowcolor[HTML]{EFEFEF} 
SAM & \multicolumn{1}{l|}{\cellcolor[HTML]{EFEFEF}CRN finetuned} & original & 76.98 & 62.70 & 76.59 & 68.89 \\ 
\xmark & \multicolumn{1}{l|}{CRN finetuned} & original & \cellcolor[HTML]{FFCE93}\textbf{78.60} & 64.74 & \cellcolor[HTML]{FFCE93}\textbf{78.18} & 69.28 \\ \hline
\end{tabular}
\vspace{-5mm}
\end{table}

Table \ref{tab:easy-hard-image-extractor} outlines the performance of two image encoders, ``CRN scratch'' and ``CRN finetuned'' (where CRN stands for Comics \resnet\;). The former is a \resnet\;trained from scratch using the self-supervised SimCLR method, and the latter, a domain-adapted \resnet\;pre-trained on ImageNet and fine-tuned with the same SimCLR approach. Original text transcriptions were used for comparison. Results suggest that models using the full panel for representation outperformed those with object-level images, indicating that a comprehensive view of the panel is advantageous for narrative understanding.

\subsection{Effect of visual encoder}
Focusing on panel-level representation, which yielded the best results in the previous experiment, we evaluated different image encoders. As we can see from Table \ref{tab:easy-hard-image-encoder}, results indicate a clear advantage of ``scratch'' and ``finetuned'' Domain Adapted \resnet\;models over the standard ImageNet-trained version. Despite being similar in performances, the ``CRN scratch'' and ``CRN finetuned'' obtain the highest accuracy in the hard and easy ood test sets. However, when compared with a bigger model such as ViT, our versions of \resnet\;hold the head with 3\% to 4\% in iid settings and above 3\% for the ood. Notably, even though \blip\;emerged as the top-performing model with its $1.3B$ total parameters in our ComicVT5 configuration, the custom models ``CRN scratch'' and ``CRN finetuned'' achieve comparable accuracies with a fifth of the total weights.

\begin{table}[]
\vspace{-5mm}
\caption{Comparison of different Image Encoders using the \vltfive\;model. The  visual representation is panel-level and the OCR transcriptions the original version. Values are reported in terms of Accuracy (\%).% \colorbox{MyOrange}{\textbf{Orange}} numbers correspond to the highest accuracy on the test-set.
}
\label{tab:easy-hard-image-encoder}
\centering
\vspace{3mm}
\begin{tabular}{rccc|cc|cc}
\multicolumn{1}{l}{} & model specifics $\downarrow$ & \multicolumn{2}{r|}{training data  $\rightarrow$} & \multicolumn{2}{c|}{\multirow{2}{*}{\textbf{easy}}} & \multicolumn{2}{c}{\multirow{2}{*}{\textbf{hard}}} \\
\multicolumn{1}{r}{\textbf{}} & \multicolumn{2}{c|}{\textbf{Image}} & \textbf{Text} & \multicolumn{2}{c|}{} & \multicolumn{2}{c}{} \\ \cline{2-8} 
\multicolumn{1}{r|}{\textbf{\begin{tabular}[c]{@{}r@{}}total\\ params\end{tabular}}} & \textbf{Extractor} & \multicolumn{1}{c|}{\textbf{Encoder}} & \textbf{\begin{tabular}[c]{@{}c@{}}Extractor\\ (OCR)\end{tabular}} & \textbf{\begin{tabular}[c]{@{}c@{}}easy\\ (iid)\end{tabular}} & \textbf{\begin{tabular}[c]{@{}c@{}}hard\\ (ood)\end{tabular}} & \textbf{\begin{tabular}[c]{@{}c@{}}easy\\ (ood)\end{tabular}} & \textbf{\begin{tabular}[c]{@{}c@{}}hard\\ (iid)\end{tabular}} \\ \cline{2-8} 
250M & \xmark & \multicolumn{1}{c|}{\resnet} & original & 74.99 & 59.99 & 69.98 & 62.45 \\ 
250M & \xmark & \multicolumn{1}{c|}{CRN scratch} & original & 78.55 & \cellcolor[HTML]{FFCE93}\textbf{64.84} & 77.00 & 69.91 \\
250M & \xmark & \multicolumn{1}{c|}{CRN finetuned} & original & 78.60 & 64.74 & \cellcolor[HTML]{FFCE93}\textbf{78.18} & 69.28 \\ 
{313M} & \xmark & \multicolumn{1}{c|}{ViT-b-16} & original & 74.34 & 58.56 & 74.60 & 64.97 \\
1.3B & \xmark & \multicolumn{1}{c|}{\blip} & original & \cellcolor[HTML]{FFCE93}\textbf{79.44} & 64.08 & 78.63 & \cellcolor[HTML]{FFCE93}\textbf{70.38} \\
\cline{2-8} 
\end{tabular}
\vspace{-5mm}
\end{table}

\subsection{Effect of text extractor}
The quality of the OCR-generated text is a crucial element in our \textcloze\;task, with poor text quality potentially leading to narrative misinterpretations. An analysis of the original OCR output revealed numerous errors, such as frequent misinterpretations of the word "amsmet" in various contexts, and minor errors like single-word mistakes or punctuation issues significantly altering the intended meaning. Consequently, we hypothesized that using updated OCR technology for context text would improve narrative understanding and model performance. Therefore, we incorporated Amazon Textract as our new generation OCR, with comparative transcription examples available in the supplementary materials.

Our investigation included a preliminary experiment without OCR context text. As Table \ref{tab:easy-hard-image-text} illustrates, the significance of OCR was evident across all image encoders. Subsequently, we assessed the impact of employing the new generation OCR. This change resulted in enhanced performance for the "easy" task with the Comics \resnet\;as the image encoder, but not with \blip\;, indicating its lesser reliance on context text. For more challenging tasks, both models showed improved performance, underscoring the heightened importance of context panel text in complex scenarios. These results, reflecting the variation in performance with different OCR inputs, underscore the critical influence of OCR quality on model effectiveness.

\begin{table}[]
\vspace{-5mm}
\caption{Comparison between no OCR (pixel-only setting), original OCR, and new generation OCR. Values are reported in terms of Accuracy (\%). \colorbox{MyGray}{Gray rows} correspond to new generation OCR.}
\label{tab:easy-hard-image-text}
\centering
\vspace{3mm}
\begin{tabular}{ccc|cc|cc}
model specifics $\downarrow$ & \multicolumn{2}{r|}{training data  $\rightarrow$} & \multicolumn{2}{c|}{} & \multicolumn{2}{c}{} \\
\multicolumn{2}{c|}{\textbf{Image}} & \textbf{Text} & \multicolumn{2}{c|}{\multirow{-2}{*}{\textbf{easy}}} & \multicolumn{2}{c}{\multirow{-2}{*}{\textbf{hard}}} \\ \hline
\textbf{Extractor} & \multicolumn{1}{c|}{\textbf{\begin{tabular}[c]{@{}c@{}}Encoder\\ \end{tabular}}} & \textbf{\begin{tabular}[c]{@{}c@{}}Extractor\\ (OCR)\end{tabular}} & \textbf{\begin{tabular}[l]{@{}c@{}}easy\\ (iid)\end{tabular}} & \textbf{\begin{tabular}[c]{@{}c@{}}hard\\ (ood)\end{tabular}} & \textbf{\begin{tabular}[l]{@{}c@{}}easy\\ (ood)\end{tabular}} & \textbf{\begin{tabular}[c]{@{}c@{}}hard\\ (iid)\end{tabular}} \\ \hline
\xmark & \multicolumn{1}{c|}{CRN scratch} & \xmark & 67.70 & 64.24 & 67.94 & 65.62 \\
\xmark & \multicolumn{1}{c|}{CRN scratch} & original & 78.55\; & 64.84\; & 77.00\; & 69.91\; \\
\rowcolor[HTML]{EFEFEF}
\xmark & \multicolumn{1}{c|}{CRN scratch} & textract & 79.1\textcolor{ForestGreen}{$\uparrow$}\; & 67.3\textcolor{ForestGreen}{$\uparrow$}\; & 77.94\textcolor{ForestGreen}{$\uparrow$} & 70.46\textcolor{ForestGreen}{$\uparrow$}\; \\
\hline
\xmark & \multicolumn{1}{c|}{\blip\;} & \xmark & 65.72 & 61.66 & 69.46 & 66.13 \\
\xmark & \multicolumn{1}{c|}{\blip\;} & original & 79.44 & 64.08 & 78.63 & 70.38 \\
\rowcolor[HTML]{EFEFEF}
\xmark & \multicolumn{1}{c|}{\blip\;} & textract & 78.28\textcolor{Red}{$\downarrow$}\; & 66.23\textcolor{ForestGreen}{$\uparrow$}\; & 78.63{\tiny $=$}\; & 71.31\textcolor{ForestGreen}{$\uparrow$}\; \\
\end{tabular}
\vspace{-5mm}
\end{table}

\subsection{Comparison with other methods}
Up to now, we have explored many configurations of the proposed ComicsVT5 architecture defining the top-performing model as the combination of panel-level features, \blip\;as Image Encoder, trained with the new generation OCR data (called ComicVT5-blip). The second best model, with just a fifth of the previous model's parameters, is the Comic Domain-Adapted \resnet\;as Image Encoder, trained from scratch with SimCLR framework on comics (indicated as ComicVT5-scratch). We compare our ComicVT5 models with the previous state-of-the-art model in Table \ref{table:comparisons}. As we can see from the table, we obtain up to a 10\% improvement over previous models which use original OCR and recurrent network architecture \cite{iyyer_amazing_2017}. Human benchmarks, cited from prior work \cite{iyyer_amazing_2017}, show 84\% accuracy on hard tasks and an assumed 100\% on easier ones (``trivial''), serving as an approximate upper bound.

\begin{table}[]
\vspace{-5mm}
\centering
\caption{Comparison of our models with previous state-of-the-art.}
\label{table:comparisons}
\vspace{3mm}
\begin{tabular}{lccccc}
                                                     &                                      & \multicolumn{4}{c}{training data}                                                                                                                                                                                                                                                  \\
\textbf{}                                            & \multicolumn{1}{c|}{\textbf{}}       & \multicolumn{2}{c|}{\textbf{easy}}                                                                                                                 & \multicolumn{2}{c}{\textbf{hard}}                                                                                             \\ \cline{3-6} 
\multicolumn{1}{c|}{\textbf{Models}}                 & \multicolumn{1}{c|}{\textbf{params}} & \textbf{\begin{tabular}[c]{@{}c@{}}easy\\ (iid)\end{tabular}} & \multicolumn{1}{c|}{\textbf{\begin{tabular}[c]{@{}c@{}}hard\\ (ood)\end{tabular}}} & \textbf{\begin{tabular}[c]{@{}c@{}}easy\\ (ood)\end{tabular}} & \textbf{\begin{tabular}[c]{@{}c@{}}hard\\ (iid)\end{tabular}} \\ \hline
\multicolumn{1}{l|}{random}                          & \multicolumn{1}{c|}{-}               & 33.3                                                          & \multicolumn{1}{c|}{33.3}                                                          & 33.3                                                          & 33.3                                                          \\
\multicolumn{1}{l|}{iyyer \cite{iyyer_amazing_2017}} & \multicolumn{1}{c|}{145M}            & 68.6                                                          & \multicolumn{1}{c|}{-}                                                             & -                                                             & 61.0                                                          \\
\multicolumn{1}{l|}{ComicVT5-scratch}                & \multicolumn{1}{c|}{250M}            & \cellcolor[HTML]{FFCE93}\textbf{79.1}                         & \multicolumn{1}{c|}{\cellcolor[HTML]{FFCE93}\textbf{67.3}}                         & 77.94                                                         & 70.46                                                         \\
\multicolumn{1}{l|}{ComicVT5-blip}                   & \multicolumn{1}{c|}{1.3B}            & 78.28                                                         & \multicolumn{1}{c|}{66.23}                                                         & \cellcolor[HTML]{FFCE93}\textbf{78.63}                        & \cellcolor[HTML]{FFCE93}\textbf{71.31}                        \\ \hline
\multicolumn{1}{l|}{human*}                          & \multicolumn{1}{c|}{-}               & 100                                                           & \multicolumn{1}{c|}{84}                                                            & 100                                                           & 84                                                           
\end{tabular}
\vspace{-3mm}
\end{table}

\subsection{Qualitative Analysis}
\begin{figure}[h]
    \centering
    \includegraphics[width=\textwidth]{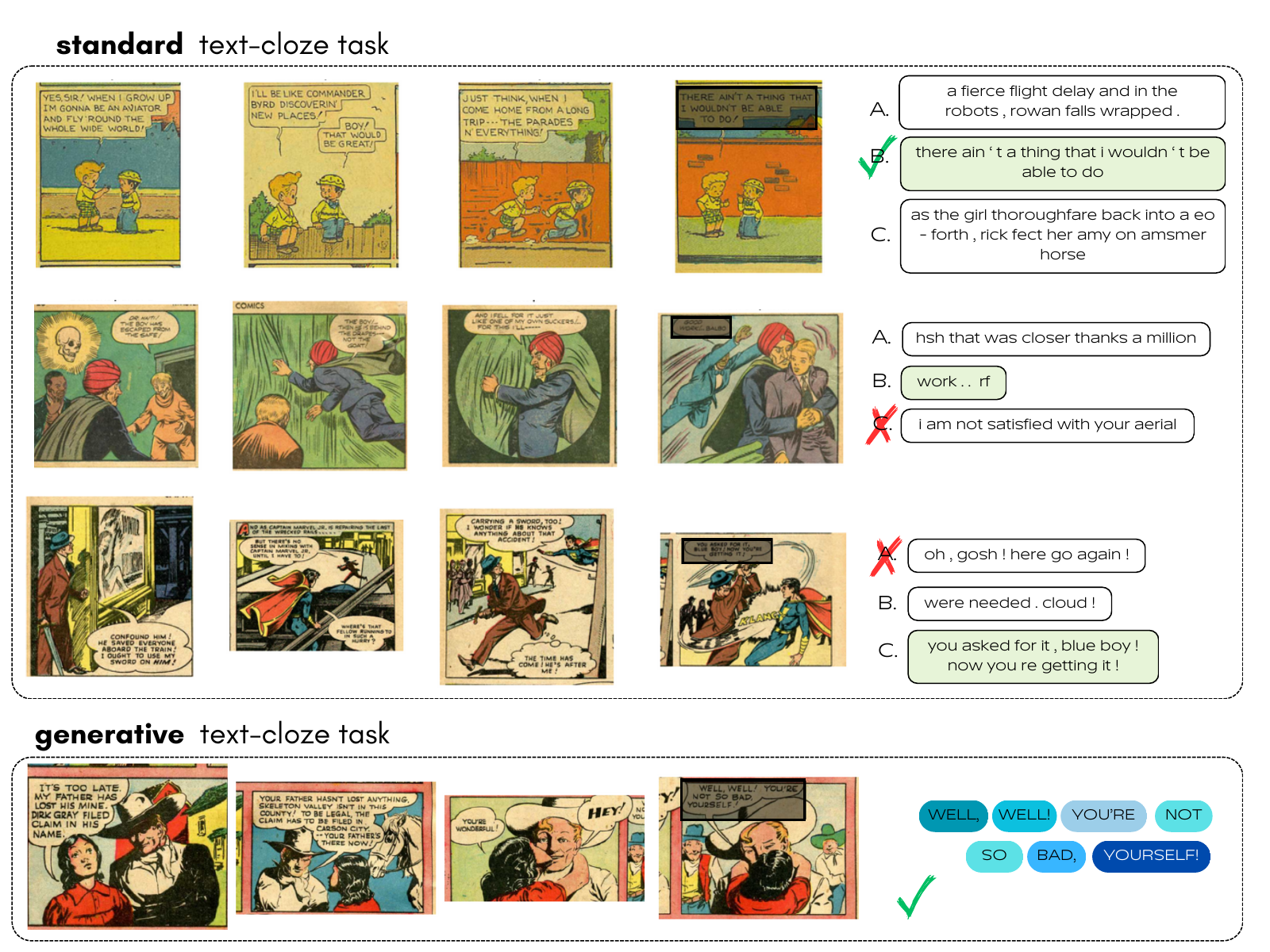}
    \caption{Examples of our ComicVT5-blip model. For every row: (i) context (first three panels), (ii) query panel (fourth panel) and (iii) options (on the right). For the \textit{generative text-cloze task} the model is trained with text generation objective.}
    \label{fig:combined_example}
    \vspace{-5mm}
\end{figure}

Our qualitative analysis examines the model's predictions for both easy and hard tasks. Figure \ref{fig:combined_example} showcases three scenarios, each illustrating a different aspect of our model's capabilities. The figure includes the context panels used by the model (the first three), the query panel with its text masked (a grey translucent box is used for visualization instead of an actual black box), and the possible text choices (with the correct answer in green). The model's selection is represented by {\color{Red} \xmark} (error) and {\color{ForestGreen} \cmark} (correct).

In the first instance depicted in Figure \ref{fig:combined_example}, our model accurately navigates a ``hard'' task sample, successfully identifying the correct narrative continuation amidst distractors sourced from adjacent pages. The first distractor is rendered implausible by its narrative incoherence, notably the incongruous inclusion of "robots", while the third distractor's lack of logical sentence structure and irrelevant mention of "girl" renders it infeasible.  The second instance in the figure exemplifies a prevalent error trend stemming from OCR inaccuracies, where flawed or incomplete transcriptions significantly alter the intended meaning, leading to erroneous selections by the model. The third instance showcases the model's proficiency in generating plausible, though not exact, narrative continuations, as evidenced by a response that, while divergent from the original text, maintains contextual relevance. This highlights the model’s capability to generate contextually coherent but not necessarily accurate predictions.

These examples collectively demonstrate the various factors influencing the performance of our model, including the accuracy of OCR transcriptions and the model’s contextual understanding in the \textcloze\;task.

\subsubsection{Further Experimental Details}
Additional exploratory studies, including those focusing on encoder-only architectures and the incorporation of increased distractors, have been detailed in the supplementary material due to space constraints in the main paper.

\subsection{Dialogues generation}
To add complexity and realism to the task, we propose a variation of \textcloze\;which involves dialogue generation. The new task aims at generating the target dialogue either from the three possible sentences provided as input or without any option provided. 

\subsubsection{With provided options.} For evaluating model performance, we utilized various natural language processing metrics, reported in the supplementary material. At inference time, for text generation, we employed beam search with 2 beams and a limit of 35 tokens, avoiding random sampling since the target output has been used as input of the model. To facilitate comparison with earlier methods we also measured accuracy, as shown in Table \ref{tab:accuracy_gen}. To calculate the accuracy, we identified the model prediction among the three options that most closely matched the meaning of the generated text. More in detail: (i) we generated embeddings for each sentence using a transformer encoder\footnote{https://huggingface.co/sentence-transformers/all-mpnet-base-v1} (a model fine-tuned on 1 billion sentence pairs), which provides contextualized embeddings for each sentence token; (ii) we compute a singular embedding for each sentence through average pooling; (iii) we compared these embeddings (generated one and the three options) using cosine similarity; (iv) the most similar option sentence is chosen for calculating accuracy. Qualitatively, as reported in the fourth row of Figure \ref{fig:combined_example}, it's apparent that the model has learned to generate %one of the three proposed 
the correct answers with minor modifications, such as eliminating extra white spaces before special characters (e.g., \textit{!}, \textit{?}).

\begin{table}[]
\centering
\caption{Comparison of different \vltfive\;Multimodal architecture. The ``Input options'' column specifies whether the model gets the three options as input. The ``Output task'' can be either classification (cls) or generation (gen). \colorbox{MyGray}{Gray rows} correspond to generative models whose accuracy has been calculated with sentence similarity.}
\label{tab:accuracy_gen}
\vspace{-2mm}
\begin{tabular}{lcccccc}
\multicolumn{1}{c}{}                                          &                                                                  &                                                                                     & \multicolumn{4}{c}{training data}                                                                                                                                                                                                                                                  \\
\multicolumn{1}{c}{}                                          &                                                                  & \multicolumn{1}{c|}{}                                                               & \multicolumn{2}{c|}{\textbf{easy}}                                                                                                                 & \multicolumn{2}{c}{\textbf{hard}}                                                                                             \\ \cline{4-7} 
\multicolumn{1}{c|}{\textbf{Models}}                          & \textbf{\begin{tabular}[c]{@{}c@{}}Input\\ options\end{tabular}} & \multicolumn{1}{c|}{\textbf{\begin{tabular}[c]{@{}c@{}}Output\\ task\end{tabular}}} & \textbf{\begin{tabular}[c]{@{}c@{}}easy\\ (iid)\end{tabular}} & \multicolumn{1}{c|}{\textbf{\begin{tabular}[c]{@{}c@{}}hard\\ (ood)\end{tabular}}} & \textbf{\begin{tabular}[c]{@{}c@{}}easy\\ (ood)\end{tabular}} & \textbf{\begin{tabular}[c]{@{}c@{}}hard\\ (iid)\end{tabular}} \\ \hline
\multicolumn{1}{l|}{ComicVT5-scratch}                         & \cmark                                                           & \multicolumn{1}{c|}{cls}                                                            & \cellcolor[HTML]{FFCE93}\textbf{79.1}                         & \multicolumn{1}{c|}{\cellcolor[HTML]{FFCE93}\textbf{67.35}}                        & 77.04                                                         & 70.46                                                         \\
\rowcolor[HTML]{EFEFEF} 
\multicolumn{1}{l|}{\cellcolor[HTML]{EFEFEF}ComicVT5-scratch} & \cmark                                                           & \multicolumn{1}{c|}{\cellcolor[HTML]{EFEFEF}gen}                                    & 72.84                                                         & \multicolumn{1}{c|}{\cellcolor[HTML]{EFEFEF}62.12}                                 & 72.95                                                         & 66.32                                                         \\
\rowcolor[HTML]{EFEFEF} 
\multicolumn{1}{l|}{\cellcolor[HTML]{EFEFEF}ComicVT5-scratch} & \xmark                                                           & \multicolumn{1}{c|}{\cellcolor[HTML]{EFEFEF}gen}                                    & 48.12                                                        & \multicolumn{1}{c|}{\cellcolor[HTML]{EFEFEF}44.80}                                 & 48.12                                                         & 44.80                                                        \\ \hline
\multicolumn{1}{l|}{ComicVT5-blip}                            & \cmark                                                           & \multicolumn{1}{c|}{cls}                                                            & 78.28                                                         & \multicolumn{1}{c|}{66.23}                                                         & \cellcolor[HTML]{FFCE93}\textbf{78.63}                        & \cellcolor[HTML]{FFCE93}\textbf{71.31}                        \\
\rowcolor[HTML]{EFEFEF} 
\multicolumn{1}{l|}{\cellcolor[HTML]{EFEFEF}ComicVT5-blip}    & \cmark                                                           & \multicolumn{1}{c|}{\cellcolor[HTML]{EFEFEF}gen}                                    & 74.58                                                         & \multicolumn{1}{c|}{\cellcolor[HTML]{EFEFEF}62.14}                                 & 73.33                                                         & 67.06                                                         \\
\rowcolor[HTML]{EFEFEF} 
\multicolumn{1}{l|}{\cellcolor[HTML]{EFEFEF}ComicVT5-blip}    & \xmark                                                           & \multicolumn{1}{c|}{\cellcolor[HTML]{EFEFEF}gen}                                    & 49.93                                                        & \multicolumn{1}{c|}{\cellcolor[HTML]{EFEFEF}46.77}                                 & 49.93                                                         & 46.77                                                       
\end{tabular}
\vspace{-5mm}
\end{table}

\subsubsection{Without input options.} In our exploration of more sophisticated dialogue generation for comics, we experimented without predefined answer options, focusing solely on context text and panel images. Details and examples of the model-generated dialogues are provided in the supplementary material.
Evaluating this approach required a shift from traditional natural language metrics due to the challenge of having only a single reference dialogue. To address this, we assessed the models using accuracy, similar to our previous methods and also experimented with feeding different proportions of the dialogue's starting tokens into the transformer decoder. For text generation, we utilized sampling with a temperature of 0.6 and a maximum length of 35 tokens.
Results indicate a decline in accuracy compared to our previous dialogue generation experiments, reflecting the increased complexity of this task. 
Finally, we evaluated NLP-specific metrics for the generation task (Rouge, Bleu, Meteor, and Rouge-l) in both generation settings (with and without input options). The detailed results of these experiments, including the variations in performance with different percentages of target decoder tokens, are presented in the supplementary material.

\section{Conclusions}
In this paper, we introduced a novel Multimodal Large Language Model architecture (Multimodal-LLM) tailored for the comics \textcloze\;task. Our investigation spanned various image representations and OCR technologies. We established that a self-supervised domain adaptation of the ResNet architecture to comic imagery achieves performance on par with recent Multimodal LLM image encoders, with significantly fewer parameters. This Domain-Adapted image encoder not only achieves state-of-the-art results in the encoder-decoder framework but also excels as a standalone component. Furthermore, we identified that the conventional \textcloze\;task, also thanks to accurate OCR, can be relatively straightforward. To address this, we designed more challenging variants of the task by introducing a generative approach. Our extensive results and ablations contribute valuable benchmarks for advanced prediction tasks in this field.

To foster further research and facilitate the development of novel architectures, we are releasing our comprehensive OCR dataset and model code, underlining our commitment to research reproducibility and innovation in comic-based language modeling.
\bibliographystyle{splncs04}
\bibliography{biblio}

\newpage

\title{Multimodal Transformer for Comics Text-Cloze}
% Second Author\inst{2,3}\orcidID{1111-2222-3333-4444}
\author{Supplementary Materials}
\authorrunning{E. Vivoli et al.}
\institute{}
\maketitle               % typeset the header of the contribution
\section*{Structure overview}
This supplementary material complements our main paper by delving deeper into specific aspects of our research that were not extensively covered due to length constraints. The content is organized as follows: in Section \ref{sec:comic-resnet} we present a nuanced analysis of the Comics domain-adapted ResNet, including t-SNE visualizations and a detailed exploration of some clusters. In Section \ref{sec:ocr}, we illustrate a comparison between the original OCR and the advanced new-generation OCR used in our study. In Section \ref{sec:ablation}, we present two different experiment results: (i) an investigation of the impact of integrating more distractors during the training phase; and (ii) the assessment of performance differences between the encoder-only and encoder-decoder versions of our architecture. Finally, in Section \ref{sec:task-variation} we provide an in-depth analysis (quantitative and qualitative results) of the proposed generative \textit{text-cloze} task.

\section{Analysis of Comics Domain-Adapted ResNet}
\label{sec:comic-resnet}

\begin{figure}[h!]
\centering
\includegraphics[width=0.7\textwidth]{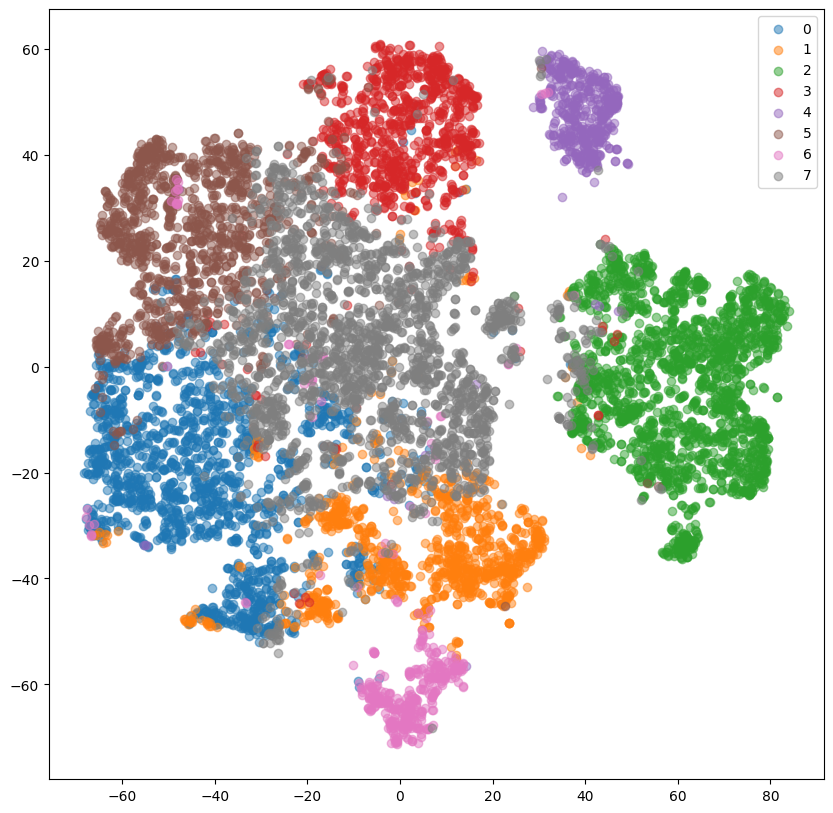}
\caption{Reduced dimensional distribution of test objects via t-SNE on Comics ResNet, featuring K-means detected clusters.}
\label{fig:tsne}
\end{figure}
\begin{figure}[h!]
\centering
\includegraphics[width=0.9\textwidth]{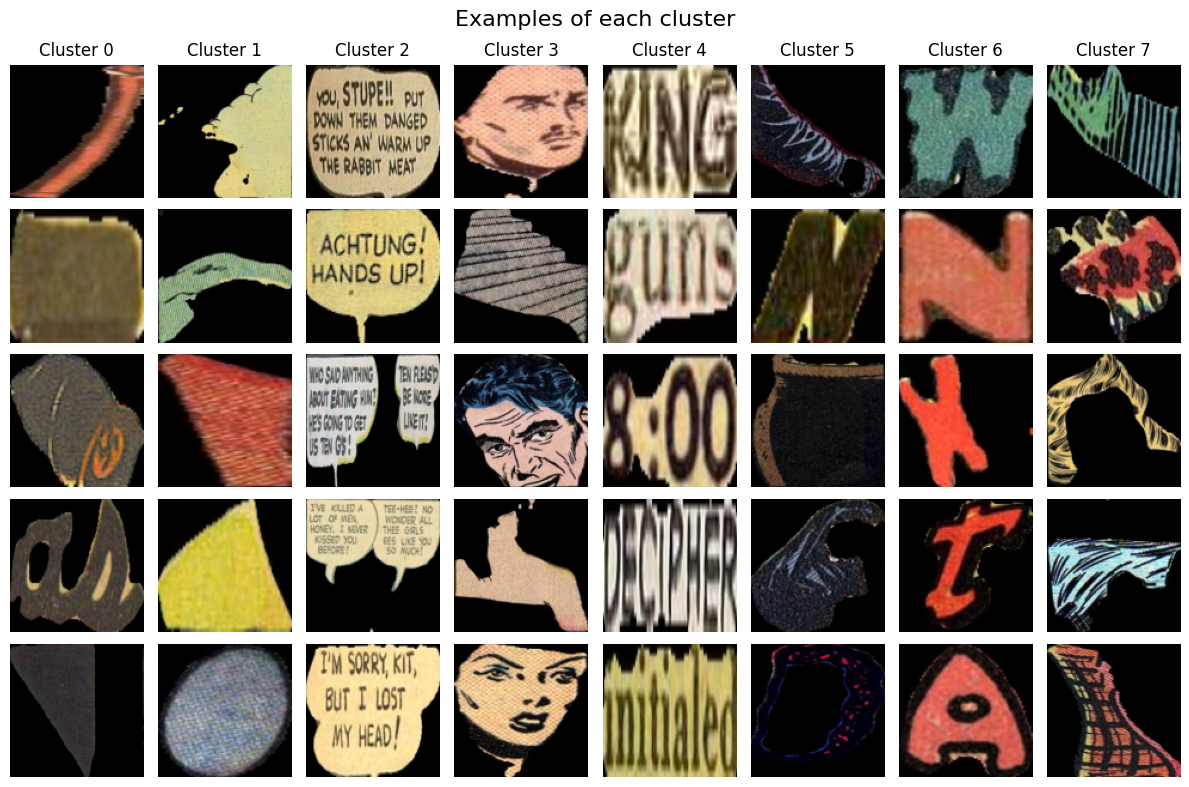}
\caption{Representative objects from each identified cluster in Figure \ref{fig:tsne}, maintaining consistent cluster labeling.}
\label{fig:examples_clusters}
\end{figure}

In Figure \ref{fig:tsne}, we present a visual representation of 10,000 test objects analyzed through a \resnet;trained from the ground up using SimCLR. To achieve a dimensional reduction from 2048 to 2, we employed the t-SNE algorithm. This figure also delineates 8 clusters, ascertained using the K-means clustering technique.
Figure \ref{fig:examples_clusters} displays a selection of five objects from each cluster identified in Figure \ref{fig:tsne}, adhering to the same cluster numbering as indicated in the figure's legend. Clusters 2, 4, and 6 predominantly include segmented balloons, words, and letters, respectively. While Cluster 3 exhibits a less pronounced trend, it predominantly features faces. The other clusters group objects with visual similarities, though they lack the distinct object type trends observed in the earlier mentioned clusters.

\section{Comparison of OCR Performance}
\label{sec:ocr}
\begin{figure}[h!]
\centering
\includegraphics[width=\textwidth]{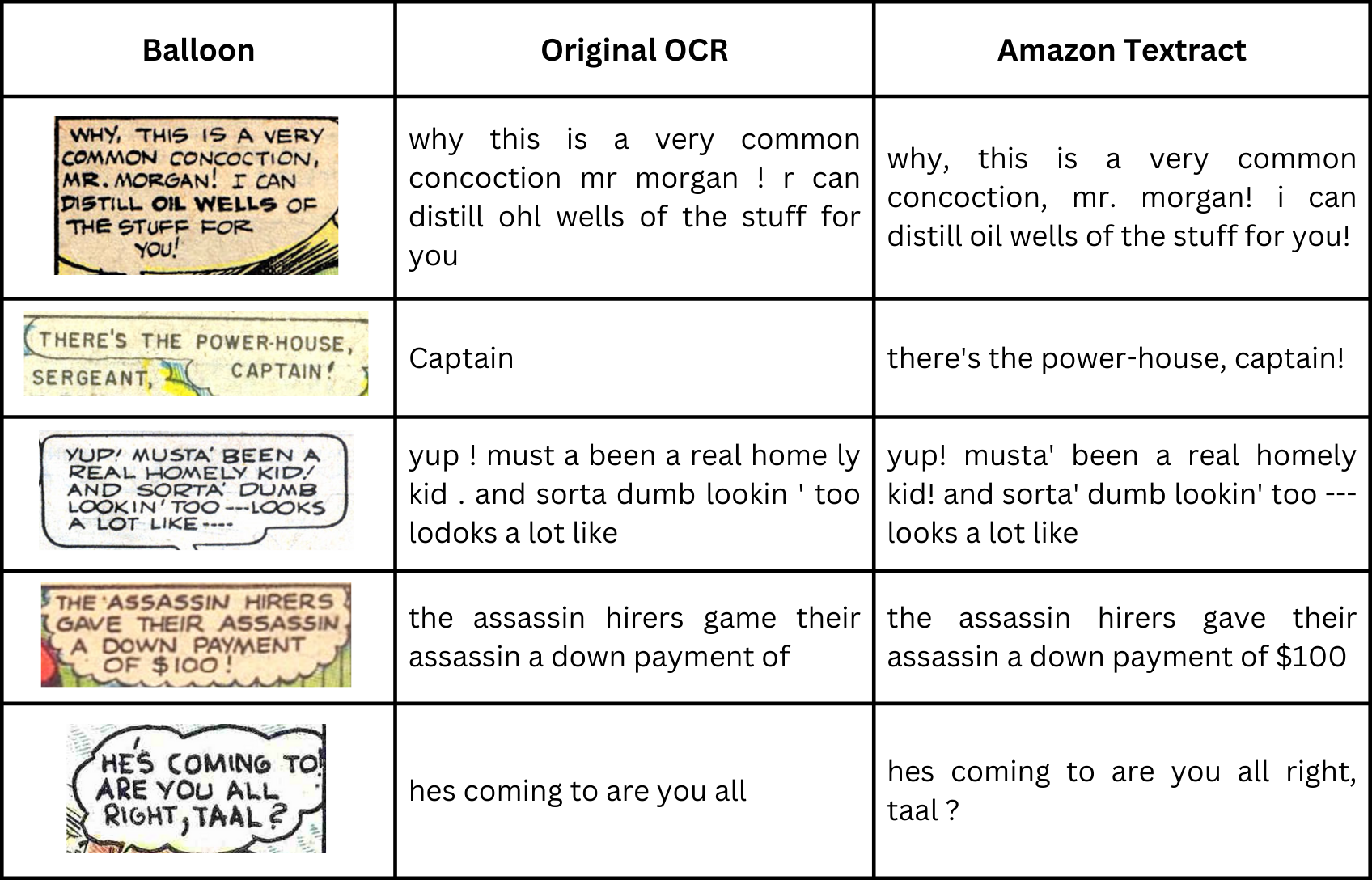}
\caption{Contrast between the original OCR and Amazon Textract transcriptions.}
\label{fig:OCR_examples}
\end{figure}

Figure \ref{fig:OCR_examples} showcases examples contrasting the original balloon image with transcriptions from two different OCR technologies. The comparison clearly demonstrates the superior quality of the new generation OCR (shown on the right).

\section{Ablation Studies}
\label{sec:ablation}

\subsection{Exploring Number of distractors}
\label{sec:number-distractors}
In an effort to increase the task's complexity, we introduced a revision of the task where the number of distractor choices was expanded from two to nine. This modification aimed to create a more challenging environment for the model's decision-making process. We constructed two variants: an ``easy'' version with distractors chosen from the entire comic dataset, and a ``hard'' version with distractors selected from pages near the target, heightening the presence of contextually significant distractors. These tasks' versions follow the same recipe as the original sets. The model's performance in this heightened complexity is detailed in Table \ref{tab1e_results_ten_options}.

\begin{table}[h!]
\caption{Model performance when trained and tested with ten distractors.}
\label{tab1e_results_ten_options}
\vspace{3mm}
\centering
\begin{tabular}{ccc|cc|cc}
model specifics $\downarrow$ & \multicolumn{2}{c|}{training data $\rightarrow$} & \multicolumn{2}{c|}{\multirow{2}{*}{\textbf{easy}}} & \multicolumn{2}{c}{\multirow{2}{*}{\textbf{hard}}} \\
\multicolumn{2}{c|}{\textbf{Image}} & \textbf{Text} & \multicolumn{2}{c|}{} & \multicolumn{2}{c}{} \\ \hline
\textbf{Extractor} & \multicolumn{1}{c|}{\textbf{Encoder}} & \textbf{\begin{tabular}[c]{@{}c@{}}Extractor\\ (OCR)\end{tabular}} & \textbf{\begin{tabular}[c]{@{}c@{}}easy\\ (iid)\end{tabular}} & \textbf{\begin{tabular}[c]{@{}c@{}}hard\\ (ood)\end{tabular}} & \textbf{\begin{tabular}[c]{@{}c@{}}easy\\ (ood)\end{tabular}} & \textbf{\begin{tabular}[c]{@{}c@{}}hard\\ (iid)\end{tabular}} \\ \hline
\xmark & CRN scratch & original & 73.61 & 37.10 & 67.15 & 56.26
\end{tabular}
\end{table}

Notably, the task difficulty rises with more distractors, especially within the ``hard'' setting where the model is tasked with distinguishing finer narrative details. The data reflects a significant enhancement in task complexity with more distractors, while also showing that in the ``easy'' variant, the model's performance remains relatively stable even with the expanded distractor pool.

\subsection{Comparing Architectural Impact}
\label{sec:comparing-architecture}
In our study, we investigated the performance impact of utilizing a transformer decoder compared to a Multilayer Perceptron (MLP) classification head. To this end, we modified the \vltfive\; encoder-decoder architecture by incorporating an MLP head that classifies among three options, following the average pooling of the encoder's output tokens.
Table \ref{table_results_transformer_decoder_MLP} presents the results of this ablation study.

\begin{table}[h!]
\centering
\caption{Comparison of different ComicVT5 Multimodal architecture. \colorbox{MyGray}{Gray rows} correspond to \vltfive\; encoder-decoder (standard \vltfive\;) while white rows to encoder-only architecture.}
\label{table_results_transformer_decoder_MLP}
\vspace{3mm}
\begin{tabular}{lcccccc}
\multicolumn{1}{c}{}                                          &                                                   &                                                                                                 & \multicolumn{4}{c}{training data}                                                                                                                                                                                                                                                  \\
\multicolumn{1}{c}{}                                          &                                                   & \multicolumn{1}{c|}{}                                                                           & \multicolumn{2}{c|}{\textbf{easy}}                                                                                                                 & \multicolumn{2}{c}{\textbf{hard}}                                                                                             \\ \cline{4-7} 
\multicolumn{1}{c|}{\textbf{Models}}                          & \multicolumn{1}{c|}{\textbf{Params}}              & \multicolumn{1}{c|}{\textbf{\begin{tabular}[c]{@{}c@{}}Multimodal\\ architecture\end{tabular}}} & \textbf{\begin{tabular}[c]{@{}c@{}}easy\\ (iid)\end{tabular}} & \multicolumn{1}{c|}{\textbf{\begin{tabular}[c]{@{}c@{}}hard\\ (ood)\end{tabular}}} & \textbf{\begin{tabular}[c]{@{}c@{}}easy\\ (ood)\end{tabular}} & \textbf{\begin{tabular}[c]{@{}c@{}}hard\\ (iid)\end{tabular}} \\ \hline
\multicolumn{1}{l|}{ComicVT5-scratch}                         & \multicolumn{1}{c|}{137M}                         & \multicolumn{1}{c|}{enc}                                                                        & 77.37                                                         & \multicolumn{1}{c|}{65.72}                                                         & 77.38                                                         & 69.95                                                         \\
\rowcolor[HTML]{EFEFEF} 
\multicolumn{1}{l|}{\cellcolor[HTML]{EFEFEF}ComicVT5-scratch} & \multicolumn{1}{c|}{\cellcolor[HTML]{EFEFEF}250M} & \multicolumn{1}{c|}{\cellcolor[HTML]{EFEFEF}enc-dec}                                            & \cellcolor[HTML]{FFCE93}\textbf{79.1}                         & \multicolumn{1}{c|}{\cellcolor[HTML]{FFCE93}\textbf{67.35}}                        & 77.04                                                         & 70.46                                                         \\ \hline
\multicolumn{1}{l|}{ComicVT5-blip}                            & \multicolumn{1}{c|}{1.2B}                         & \multicolumn{1}{c|}{enc}                                                                        & 76.63                                                         & \multicolumn{1}{c|}{63.61}                                                         & 77.23                                                         & 70.69                                                         \\
\rowcolor[HTML]{EFEFEF} 
\multicolumn{1}{l|}{\cellcolor[HTML]{EFEFEF}ComicVT5-blip}    & \multicolumn{1}{c|}{\cellcolor[HTML]{EFEFEF}1.3B} & \multicolumn{1}{c|}{\cellcolor[HTML]{EFEFEF}enc-dec}                                            & 78.28                                                         & \multicolumn{1}{c|}{\cellcolor[HTML]{EFEFEF}66.23}                                 & \cellcolor[HTML]{FFCE93}\textbf{78.63}                        & \cellcolor[HTML]{FFCE93}\textbf{71.31}                       
\end{tabular}
\end{table}

It reveals a marginal performance reduction when omitting the \vltfive\; decoder, despite the encoder-only variant having fewer learnable parameters. On the other hand, this also suggests the complete encoder-decoder framework offers a slight advantage, especially for the more intricate 'hard' version of the task.

\section{Generative \textit{text-cloze} Task}
\label{sec:task-variation}

\subsection{Quantitative Results}
\label{sec:generation-quantitative}

\begin{table}[h!]
\centering
 \caption{Metrics for the dialogue generation models giving the three possible answers of the \textcloze\;task in the input of the encoder.}
\label{tab:nlp_metrics_gen}
\vspace{3mm}
\begin{tabular}{lccccccc}
                                                           & \multicolumn{1}{l}{}                              & \multicolumn{4}{c}{\textbf{easy (iid)}}                                                                                                                               \\
\multicolumn{1}{c}{\textbf{Models}}                        & \multicolumn{1}{c|}{\textbf{Params}}          & \multicolumn{1}{l|}{\textbf{bleu1}}    & \multicolumn{1}{l|}{\textbf{bleu4}}     & \multicolumn{1}{l|}{\textbf{rouge1}} & \multicolumn{1}{l|}{\textbf{rouge2}}    & \multicolumn{1}{l|}{\textbf{meteor}}    & \multicolumn{1}{l}{\textbf{rouge-l}}    \\ \cline{3-8} 
\multicolumn{1}{l|}{ComicVT5-scratch}                      & \multicolumn{1}{c|}{250M}                        & 0.6543 & 0.6198                                  &  0.7216 & 0.7075                                  & 0.6699                                  & 0.7203                                  \\
\rowcolor[HTML]{FFCE93} 
\multicolumn{1}{l|}{\cellcolor[HTML]{EFEFEF}ComicVT5-blip} & \multicolumn{1}{c|}{\cellcolor[HTML]{EFEFEF}1.3B} & \textbf{0.6663} & \textbf{0.6334} & \textbf{0.7380}                        & \textbf{0.7249}                         & \textbf{0.6854}                         & \textbf{0.7370}                         \\
                                                           & \multicolumn{1}{l}{}                              & \multicolumn{4}{c}{\textbf{easy (ood)}}                                                                                                                               \\
                                                           & \multicolumn{1}{l|}{}                       & \multicolumn{1}{l|}{\textbf{bleu1}}      & \multicolumn{1}{l|}{\textbf{bleu4}}     &
                                                           \multicolumn{1}{l|}{\textbf{rouge1}} & \multicolumn{1}{l|}{\textbf{rouge2}}    & \multicolumn{1}{l|}{\textbf{meteor}}    & \multicolumn{1}{l}{\textbf{rouge-l}}    \\ \cline{3-8} 
\multicolumn{1}{l|}{ComicVT5-scratch}                      & \multicolumn{1}{c|}{250M}                         & 0.6525 & 0.6181                                & 0.7227  & 0.7085                                  & 0.6702                                  & 0.7216                                  \\
\rowcolor[HTML]{EFEFEF} 
\multicolumn{1}{l|}{\cellcolor[HTML]{EFEFEF}ComicVT5-blip} & \multicolumn{1}{c|}{\cellcolor[HTML]{EFEFEF}1.3B} & 0.6599 & 0.6249 & 0.7267                                  & 0.7120                                  & 0.6740                                  & 0.7251                                  \\ \hline
                                                           & \multicolumn{1}{l}{}                              & \multicolumn{4}{c}{\textbf{hard (ood)}}                                                                                                                               \\
                                                           & \multicolumn{1}{l|}{}                        
                                                           & \multicolumn{1}{l|}{\textbf{bleu1}} & \multicolumn{1}{l|}{\textbf{bleu4}}     & 
                                                           \multicolumn{1}{l|}{\textbf{rouge1}} & \multicolumn{1}{l|}{\textbf{rouge2}}    & \multicolumn{1}{l|}{\textbf{meteor}}    & \multicolumn{1}{l}{\textbf{rouge-l}}    \\ \cline{3-8} 
\multicolumn{1}{l|}{ComicVT5-scratch}                      & \multicolumn{1}{c|}{250M}                        & 0.5834 & 0.5439  & 0.6268                                & 0.6039                                  & 0.5808                                  & 0.6243                                  \\
\rowcolor[HTML]{EFEFEF} 
\multicolumn{1}{l|}{\cellcolor[HTML]{EFEFEF}ComicVT5-blip} & \multicolumn{1}{c|}{\cellcolor[HTML]{EFEFEF}1.3B} & 0.5718 & 0.5329      & 0.6274                            & 0.6044                                  & 0.5817                                  & 0.6246                                  \\
                                                           & \multicolumn{1}{l}{}                              & \multicolumn{4}{c}{\textbf{hard (iid)}}                                                                                                                               \\
                                                           & \multicolumn{1}{l|}{}                         
                                                           & \multicolumn{1}{l|}{\textbf{bleu1}} & \multicolumn{1}{l|}{\textbf{bleu4}}     &
                                                           \multicolumn{1}{l|}{\textbf{rouge1}} & \multicolumn{1}{l|}{\textbf{rouge2}}    & \multicolumn{1}{l|}{\textbf{meteor}}    & \multicolumn{1}{l}{\textbf{rouge-l}}    \\ \cline{3-8} 
\multicolumn{1}{l|}{ComicVT5-scratch}                      & \multicolumn{1}{c|}{250M}                        & 0.6130 & \cellcolor[HTML]{FFCE93}\textbf{0.5746} & 0.6657 & 0.6450                                  & 0.6169                                  & 0.6633                                  \\
\rowcolor[HTML]{EFEFEF} 
\multicolumn{1}{l|}{\cellcolor[HTML]{EFEFEF}ComicVT5-blip} & \multicolumn{1}{c|}{\cellcolor[HTML]{EFEFEF}1.3B} & \cellcolor[HTML]{FFCE93}\textbf{0.6131} & 0.5738 & \cellcolor[HTML]{FFCE93}\textbf{0.6721}                                 & \cellcolor[HTML]{FFCE93}\textbf{0.6520} & \cellcolor[HTML]{FFCE93}\textbf{0.6238} & \cellcolor[HTML]{FFCE93}\textbf{0.6706}
\end{tabular}
\end{table}

Table \ref{tab:nlp_metrics_gen} displays the natural language processing (NLP) metrics for dialogues generated with the \textcloze; task's three options fed into the transformer encoder. The results showcase the proficient performance of the two models trained for this task in accurately recalling the correct dialogue from the three options.

\begin{table}[]
\centering
\caption{Metrics for the dialogue generation models without seeing any of the possible answers of \textcloze\;task in the encoder input. The percentage specified for each evaluation corresponds to the percentage of the target tokens used on the transformer decoder, these tokens are not considered on the evaluation.}
\label{tab:nlp_metrics_gen_no_opt}
\vspace{3mm}
\begin{tabular}{lccccccc}
                                                           &                                                   &                                         & \multicolumn{4}{c}{\textbf{30\%}}                                                                                                                                     \\
\multicolumn{1}{c}{\textbf{Models}}                        & \multicolumn{1}{c|}{\textbf{Params}}              & \multicolumn{1}{c|}{\textbf{bleu1}}     & \multicolumn{1}{c|}{\textbf{bleu4}}  & \multicolumn{1}{c|}{\textbf{rouge1}}   & \multicolumn{1}{c|}{\textbf{rouge2}}    & \multicolumn{1}{c|}{\textbf{meteor}}    & \textbf{rouge-l}                        \\ \cline{3-8} 
\multicolumn{1}{l|}{ComicVT5-scratch}                      & \multicolumn{1}{c|}{250M}                         & 0.1600                                  & 0.0038                                  & 0.1783 & 0.0302                                  & 0.1332                                  & 0.1716                                  \\
\rowcolor[HTML]{EFEFEF} 
\multicolumn{1}{l|}{\cellcolor[HTML]{EFEFEF}ComicVT5-blip} & \multicolumn{1}{c|}{\cellcolor[HTML]{EFEFEF}1.3B} & 0.1577                                  & \cellcolor[HTML]{FFCE93}\textbf{0.0044} & 0.1777 & 0.0310                                  & 0.1336                                  & 0.1707                                  \\
                                                            &                                                   &                                         & \multicolumn{4}{c}{}                                                                                                                                     \\
                                                           &                                                   &                                         & \multicolumn{4}{c}{\textbf{50\%}}                                                                                                                                     \\
                                                           & \multicolumn{1}{c|}{}                             & \multicolumn{1}{c|}{\textbf{bleu1}}     & \multicolumn{1}{c|}{\textbf{bleu4}}     &
                                                            \multicolumn{1}{c|}{\textbf{rouge1}} &\multicolumn{1}{c|}{\textbf{rouge2}}    & \multicolumn{1}{c|}{\textbf{meteor}}    & \textbf{rouge-l}                        \\ \cline{3-8} 
\multicolumn{1}{l|}{ComicVT5-scratch}                      & \multicolumn{1}{c|}{250M}                         & 0.1717                                  & 0.0035 & 0.1983                                  & 0.0329                                  & 0.1501                                  & 0.1946                                  \\
\rowcolor[HTML]{EFEFEF} 
\multicolumn{1}{l|}{\cellcolor[HTML]{EFEFEF}ComicVT5-blip} & \multicolumn{1}{c|}{\cellcolor[HTML]{EFEFEF}1.3B} & 0.1699                                  & 0.0032 & 0.1978 &                                 0.0337                                  & 0.1491                                  & 0.1939                                  \\
                                                            &                                                   &                                         & \multicolumn{4}{c}{}                                                                                                                                     \\
                                                           &                                                   &                                         & \multicolumn{4}{c}{\textbf{75\%}}                                                                                                                                     \\
                                                           & \multicolumn{1}{c|}{}                             & \multicolumn{1}{c|}{\textbf{bleu1}}     & \multicolumn{1}{c|}{\textbf{bleu4}}     &
                                                           \multicolumn{1}{c|}{\textbf{rouge1}} & \multicolumn{1}{c|}{\textbf{rouge2}}    & \multicolumn{1}{c|}{\textbf{meteor}}    & \textbf{rouge-l}                        \\ \cline{3-8} 
\multicolumn{1}{l|}{ComicVT5-scratch}                      & \multicolumn{1}{c|}{250M}                         & \cellcolor[HTML]{FFCE93}\textbf{0.2389} & 0.0034 & \cellcolor[HTML]{FFCE93}\textbf{0.2661}                                 & 0.0422                                  & 0.2151                                  & \cellcolor[HTML]{FFCE93}\textbf{0.2646} \\
\rowcolor[HTML]{EFEFEF} 
\multicolumn{1}{l|}{\cellcolor[HTML]{EFEFEF}ComicVT5-blip} & \multicolumn{1}{c|}{\cellcolor[HTML]{EFEFEF}1.3B} & 0.2363                                  & 0.0036 & 0.2653                                 & \cellcolor[HTML]{FFCE93}\textbf{0.0454} & \cellcolor[HTML]{FFCE93}\textbf{0.2171} & 0.2638                                 
\end{tabular}
\end{table}

Similar to before, Table \ref{tab:nlp_metrics_gen_no_opt} presents the NLP metrics for dialogues generated without the inclusion of possible answers in the encoder input. The table shows results calculated based on the proportion of the target dialogue provided to the transformer decoder. The results indicate a trend where providing a larger percentage of the target dialogue leads to a sentence completion more aligned with the target itself. Interestingly, these results highlight a lower performance relative to the previous approach, reflecting the increased task complexity when no option is provided as input.

\subsection{Qualitative results}
\label{sec:generation-qualitative}
\begin{figure}
\centering
\begin{subfigure}[b]{0.9\textwidth}
   \includegraphics[width=1\linewidth]{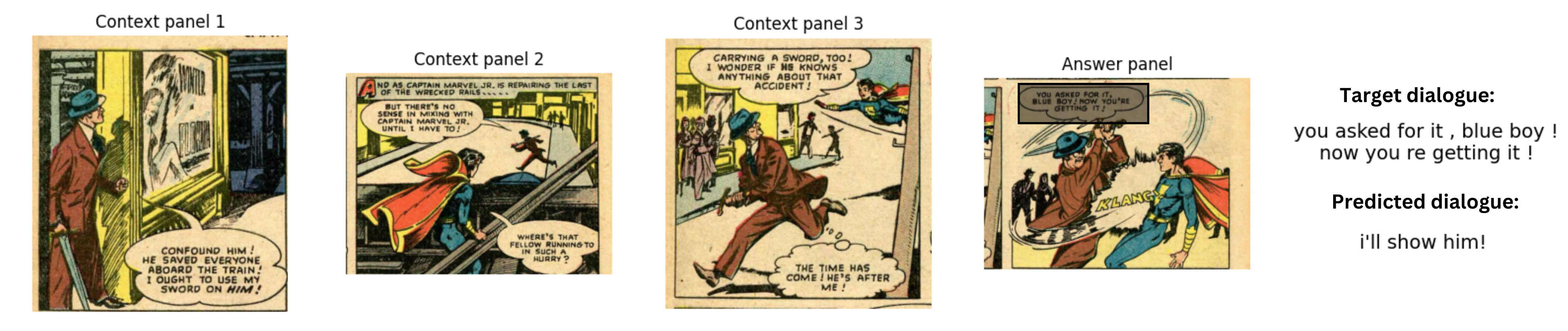}
   \caption{}
   \label{fig:Ng1} 
\end{subfigure}%
\\
\begin{subfigure}[b]{0.9\textwidth}
   \includegraphics[width=1\linewidth]{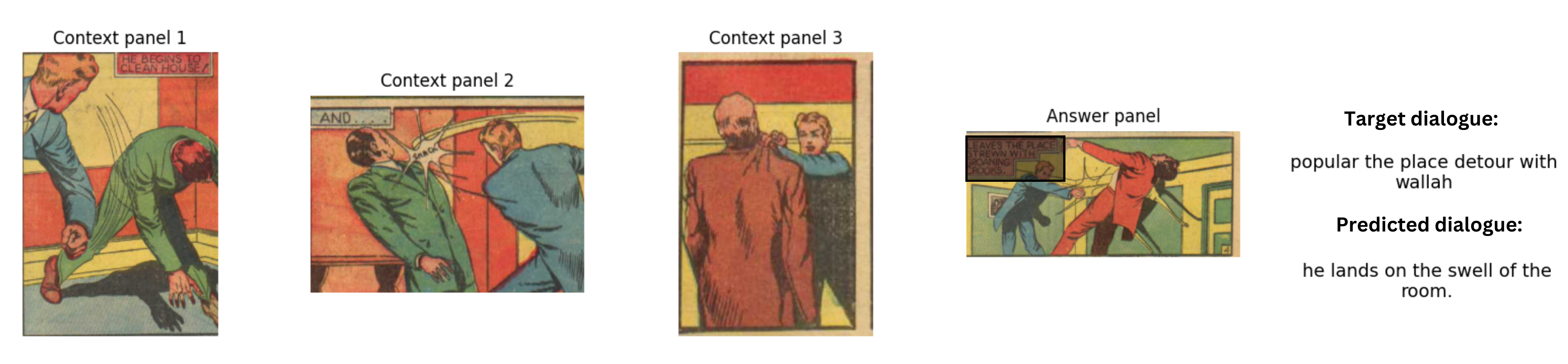}
   \caption{}
   \label{fig:Ng2} 
\end{subfigure}%
\\
\begin{subfigure}[b]{0.9\textwidth}
   \includegraphics[width=1\linewidth]{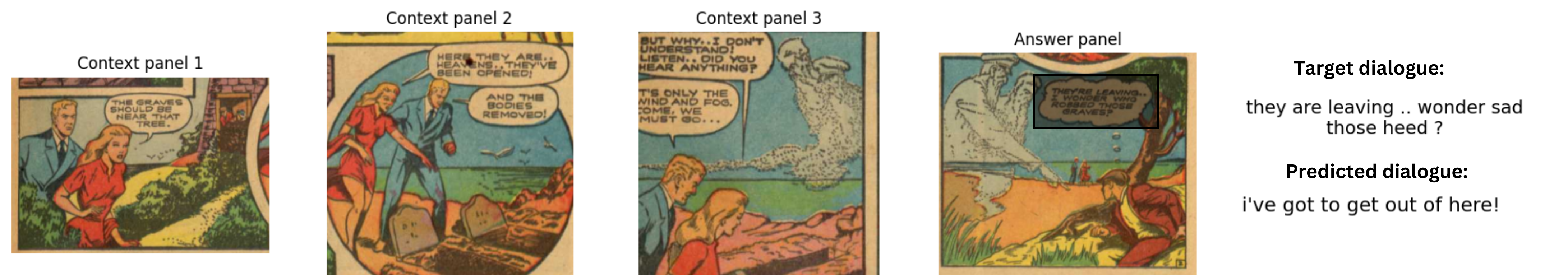}
   \caption{}
   \label{fig:Ng3}
\end{subfigure}

\caption[Three examples of generation task]{(a) Example of a contextually appropriate generated dialogue; (b) Another instance of a feasible dialogue generation; (c) An example of a repeated dialogue that does not match the current story context.}
\end{figure}

This section showcases three dialogue examples generated by the Comics-VT5-blip model, utilizing textract OCR for context text and without providing predefined answer options. The generation of these dialogues employed a beam search approach with 2 beams and a cap of 35 tokens. The examples demonstrate the model's capabilities in dialogue generation, including two instances of contextually appropriate dialogues (Figures \ref{fig:Ng1} and \ref{fig:Ng2}) and one instance where the generated dialogue, while coherent, does not align well with the ongoing narrative (Figure \ref{fig:Ng3}).

\end{document}